\documentclass{article}
\usepackage{ijcai20}

\usepackage{times}
\usepackage{soul}
\usepackage{url}
\usepackage[utf8]{inputenc}
\usepackage[small]{caption}
\usepackage{graphicx}
\usepackage{amsmath}
\usepackage{amsthm}
\usepackage{booktabs}
\usepackage{algorithm}
\usepackage{algorithmic}
\usepackage{footnote}
\usepackage{multirow}
\usepackage{color}
\title{KLDivNet: An unsupervised neural network for multi-modality image registration}

\author{
Yechong Huang$^{1,2}$, Tao Song$^2$, Jiahang Xu$^1$, Yinan Chen$^2$, Xiahai Zhuang$^1$\footnote{Contact Author}\\
\affiliations
$^1$School of Data Science, Fudan University, Shanghai, China\\
$^2$SenseTime Technology, Shanghai, China\\
\emails
zxh@fudan.edu.cn
}

\begin{document}

\maketitle

\begin{abstract}
Multi-modality image registration is one of the most underlined processes in medical image analysis.
Recently, convolutional neural networks (CNNs) have shown significant potential in deformable registration.
However, the lack of voxel-wise ground truth challenges the training of CNNs for an accurate registration.
In this work, we propose a cross-modality similarity metric, based on the KL-divergence of image variables, and implement an efficient estimation method using a CNN.
This estimation network, referred to as KLDivNet, can be trained unsupervisedly.
We then embed the KLDivNet into a registration network to achieve the unsupervised deformable registration for multi-modality images.
We employed three datasets, i.e., AAL Brain, LiTS Liver and Hospital Liver, with both the intra- and inter-modality image registration tasks for validation.
Results showed that our similarity metric was effective, and the proposed registration network delivered superior performance compared to the state-of-the-art methods. 
\textit{Codes will be released on github.}

\end{abstract}

\section{Introduction}

\begin{figure}[t]
  \centering
  \includegraphics[scale=0.5]{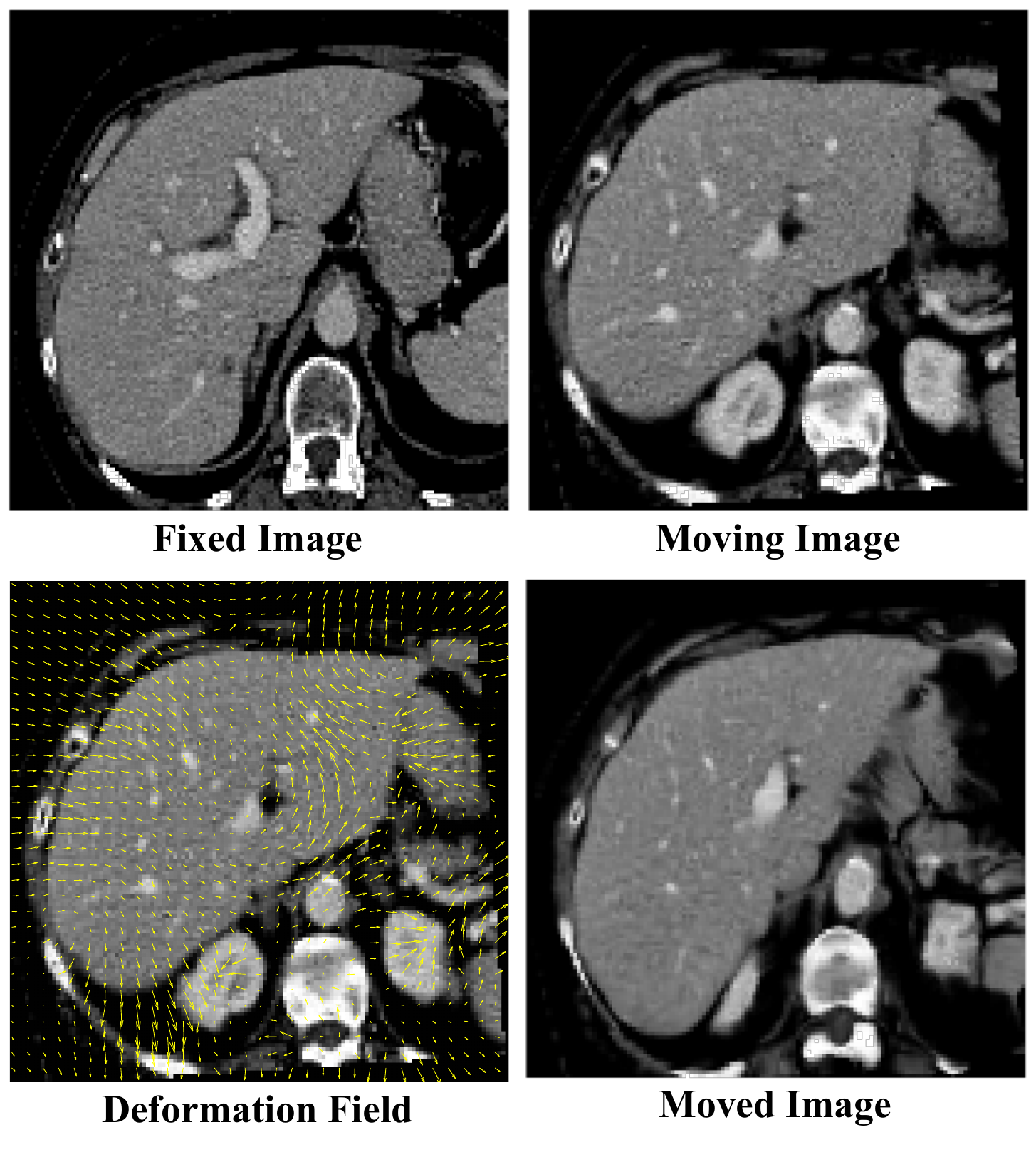}
  \caption{An example of Deformable Registration. The fixed and moving images are inter-subject CT images from the LiTS dataset \protect\cite{bilic2019liver}. The DVF is given by our registration method.}
  \label{fig:Figure1}
\end{figure}

Image registration, the process that aligns two or more images of the same scene, is a fundamental procedure in medical image analysis.
According to the types of transformations, registration can be categorised into two groups, i.e., linear registration and deformable registration \cite{journal/pmb/Hill01}.
Deformable registration allows nonuniform alignments and can be represented by a parameterized model, such as free-form deformation (FFD) \cite{FFD1999rueckert,journal/tmi/Zhuang11}, or a nonparametric displacement vector field (DVF) \cite{LDDMM2005,elasticReg1999}.
An example of deformable registration is shown in Figure \ref{fig:Figure1}.

In general, the idea of learning-based registration framework refers to the deep neural networks (DNNs) which input a moving image and a fixed image and output (predict) transform parameters \cite{de2019deep,haskins2019deep,VoxelMorph2018,VoxelMorph2019}.
In the training stage, the loss function can be set to an image similarity metric to obtain an unsupervised learning.
The choice of similarity metrics is of great significance for the learning-based methods.
In intra-modality registration, intensity difference-based metrics, such as the sum of squared differences (SSD) and cross correlation (CC), can be used \cite{de2019deep}.
However, in the inter-modality registration the intensities of the two images are typically not linearly correlated, and the conventional intensity difference-based metrics may not correctly reflect the similarity of images.

Information theoretic metrics, such as mutual information (MI) and normalized MI, were developed for multi-modality image registration \cite{viola1997alignment,maes1997multimodality,journal/pr/Studholme99}.
MI quantifies the mutual dependencies of intensity-pairs and is robust in the multi-modality situation.
However, conventionally the calculation of MI needs to compute the joint distribution, via joint histogram of images, which can be arduous and inefficient via DNNs.
Particularly, the calculation of MI gradient, attributed to the gradient of joint histogram, with the back-propagation (BP) scheme is challenging in the deep-learning architecture.


In this work, we propose to compute the multi-modality image similarity via the Kullback-Leibler (KL-) divergence \cite{KLDiv1997} and a DNN-based estimator.
This is inspired by Belghazi {\it et al.}\shortcite{MINE2018}, who proposed a new idea of computing mutual information using DNNs, i.e., the mutual information neural estimation (MINE).
The DNN estimator of the similarity metric, referred to as KLDivNet, is then used as a loss function and embedded into a learning-based registration framework to achieve a registration network. 

This registration network, denoted as DivRegNet, is designed for multi-modality registration thanks to the multi-modality similarity loss function of KLDivNet.
Furthermore, DivRegNet is an unsupervised learning scheme, since KLDivNet is an intensity-based similarity, which only takes the input of the fixed and moved images and can be trained unsupervisedly.

The contributions of this work are listed as follows: (1) we present a KLDivNet module to estimate the multi-modality image similarity, which can be training unsupervisedly; (2) we propose an unsupervised registration network based on KLDivNet for multi-modality image registration; (3) we demonstrate the proposed framework with state-of-the-art performance using three registration tasks.

\begin{figure*}[t]
  \centering
  \includegraphics[scale=0.55]{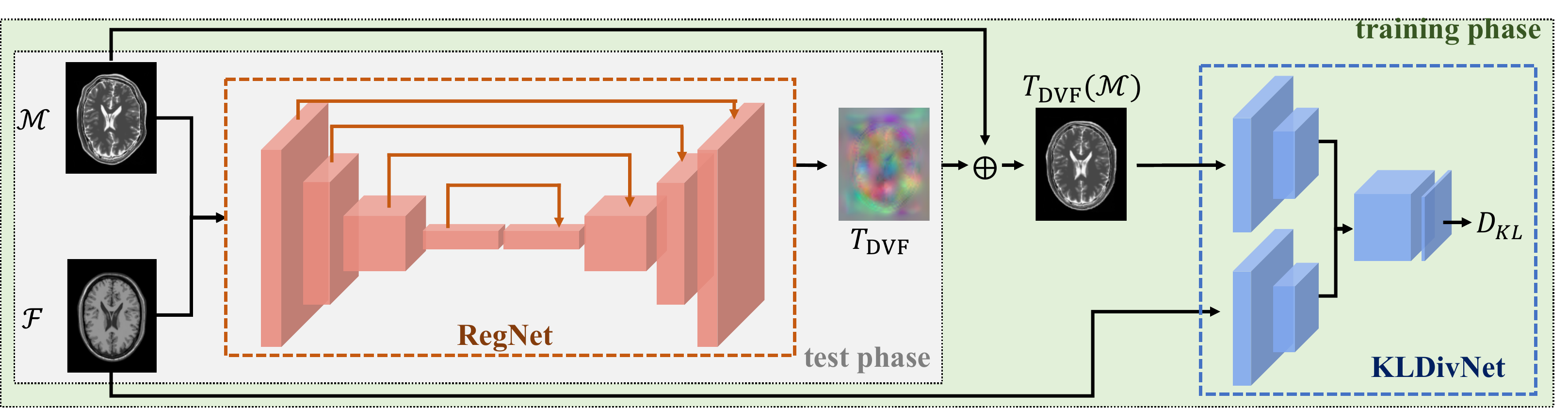}
  \caption{The framework and architecture of DivRegNet, which consists of a registration network (RegNet) to perform the registration and KLDivNet to estimate multi-modality image similarity. In the training phase, the framework is optimized in one-stage; in the inference (test) phase, only the RegNet is used.}
  \label{fig:Figure2}
\end{figure*}

\section{Related Works}

The development of deep learning presents a good opportunity to solve the problem of unsupervised deformation registration. Recently, several learning-based image registration methods have been proposed with different architecture \cite{haskins2019deep}.
Balakrishnan {\it et al.}\shortcite{VoxelMorph2018,VoxelMorph2019} adopted a Unet-like structure to generate the dense deformation field in the registration, called VoxelMorph. In this structure, a parametrized registration function is learned from a collection of volumes by CNN. The optimization of the registration parameter is achieved by evaluating the learned function on the given volumes, resulting in rapid registration.
Shen {\it et al.}\shortcite{JointAffine2019} proposed an end-to-end deep-learning framework combining an affine registration and a deformation registration. After trained, the proposed method can complete the registration process in one forward pass.
Zhao {\it et al.}\shortcite{Recursivecascaded2019} presented a recursive cascaded network for unsupervised deformable image registration. By the proposed architecture, moving image is warped by each cascade scheme and finally aligned to the fixed image. Next, every cascade learns to perform a progressive deformation by the recursive scheme.
In the mentioned four works, localized normalized cross correlation (LNCC) is employed as a main part of loss function. However, LNCC loss has a significant drawback, as it is incapable to handle the multi-modality registration scheme.

Another topic related to our work is using neural network as a similarity metric. Recently, Belghazi {\it et al.}\shortcite{MINE2018} proposed a mutual information neural estimator (MINE) to estimate the MI between high dimensional continuous random variables. By MINE, the value of MI can be achieved by gradient descent over neural networks. After MINE is first proposed, various works applied MINE as loss function to for different tasks, such as unsupervised representation learning \cite{MaxInfoMINE2018} and entropy estimator analysis \cite{FormalMINE2018}. Our work followed MINE in this regard, and applied the estimation of KL-divergence into the unsupervised deformation registration problem.

\section{DivRegNet for unsupervised multi-modality image registration}

\subsection{KLDivNet: Estimation of image similarity via KL-divergence and neural networks}
\label{sec:KLDivNet}

In this section, we illustrate the core idea of KLDivNet.
Given random variables $\mu$ and $\lambda$, KL divergency of these two variables is computed as follows,
\begin{align}
\label{equation:2}
    D_{KL}(\mu \Vert \lambda) =  \int_{x} \log \frac{\mu(x)}{\lambda(x)} \mu(\mathrm{d} x)
    = E_\mu \log \frac{\mu(X)}{\lambda(X)} . 
\end{align}
For two random variables of image $I_F$ and $I_M$,  MI of them is given by,
\begin{align}
\label{equation:1}
    \mathrm{MI}(I_\mathcal{F},I_\mathcal{M}) =  \int_{i_{_\mathcal{F}}, i_{_\mathcal{M}}} \!\!\!\!\!\!p(i_{_\mathcal{F}}, i_{_\mathcal{M}}) \log \frac{p(i_{_\mathcal{F}}, i_{_\mathcal{M}})}{p(i_{_\mathcal{F}})p(i_{_\mathcal{M}})} \mathrm{d} i_{_\mathcal{F}} \mathrm{d} i_{_\mathcal{M}}
\end{align}
By considering the joint distribution and the product of marginal distributions of $I_\mathcal{F}$ and $I_M$ as $\mu$ and $\lambda$, respectively, one can see that MI of image $I_\mathcal{F}$ and $I_\mathcal{M}$ is equivalent to a KL-divergence form, as follows,
\begin{align}
\label{equation:1_2}
     D_{KL}(P_{I_\mathcal{F},I_\mathcal{M}} \Vert P_{I_\mathcal{F}} P_{I_\mathcal{M}}) = \mathrm{MI}(I_\mathcal{F},I_\mathcal{M}),
\end{align}
where, $P_{I_\mathcal{F},I_\mathcal{M}}$ is the joint distribution, and $P_{I_\mathcal{F}} P_{I_\mathcal{M}}$ is the product of marginal distributions.

KL-divergence has fine properties, such as nonnegativity, while MI has been widely proved to be applicable to multi-modality image registration.
In the following, we deduce a similarity metric by applying the Donsker-Varadhan lower bound of KL-divergence \cite{donskerIII1976,donskerIV1983}.
We show an efficient estimation of the similarity metric via an optimization problem based on a deep learning network, i.e., the KLDivNet,
and the optimization problem is solved by maximizing a bounded above objective function.


\subsubsection{Donsker-Varadhan representation of KL-divergence}
\label{sec:DV-representation}

According to the Donsker-Varadhan variational representation, with the probability measures, $\mu$ and $\lambda$, and a space of bounded measurable functions, $\mathcal{B}(x)$, on a measurable space $(X,\Sigma)$, for a given probability distribution function $u(x)$ the KL-divergence can be written as follows,
\begin{align}
\label{equation:3}
	& D_{KL}(\mu \Vert \lambda) \nonumber \\
	&= \int_x \log \frac{\mu(x)}{\lambda(x)} \cdot \frac{\int_x u(x)\lambda(d x)}{u(x)} \cdot \frac{u(x)}{\int_x u(x)\lambda(d x)} \mu(d x) \nonumber \\
	&= \int_x \log \frac{\mu(x)}{\lambda(x)} \cdot \frac{\int_x u(x)\lambda(d x)}{u(x)} \mu(d x) \nonumber \\
	& \quad + \int_x \log \frac{u(x)}{\int_x u(x)\lambda(d x)} \mu(d x) \nonumber \\
    &=\mathrm{Comp}_1+\mathrm{Comp}_2.
\end{align}
Let $z(x)=\frac{\lambda(x) \cdot u(x)}{\int_x u(x)\lambda(d x)}$, which is a probability distribution function of $x$. The first component in Eq. (\ref{equation:3}) becomes,
\begin{align}
\label{equation:4}
	\mathrm{Comp}_1 = \int_x \log \frac{\mu(x)}{z(x)} \mu(d x) = D_{KL}(\mu \Vert z) \geq 0 .
\end{align}%
Referring to Eq. (\ref{equation:3}), we have
\begin{align}
\label{equation:5}
	D_{KL}(\mu \Vert \lambda) \geq& \int_x \log \frac{u(x)}{\int_x u(x)\lambda(d x)} \mu(d x) \nonumber \\
	=& \int_x \log u(x) \mu(d x) - \log \int_x u(x)\lambda(d x) .
\end{align}%
In particular, let $u(x)=e^{\Phi}$, where $\Phi$ is any function in the space $\mathcal{B}(x)$, we have the upper bound form,
\begin{align}
\label{equation:6}
D_{KL}(\mu \Vert \lambda) \geq \sup_{\Phi} E_{\mu}[\Phi] - \log E_{\lambda}[e^{\Phi}] =S.
\end{align}%
This upper bound is tight, as Eq. (\ref{equation:6}) holds the equality with $\Phi^\star=\log [\mu(x)/\lambda(x)] + c$, where $c$ is a constant.

We propose to substitute the maximum of the lower bound for  the maximum of KL-divergence (or MI) in image registration.
Hence, the lower bound, i.e., the right-hand-side term $S$ of Eq. (\ref{equation:6}), can be considered as a similarity when the variables are images.

\subsubsection{Estimation of image similarity using KLDivNet}

We denote the moving and fixed images as $\mathcal{M}$ and $\mathcal{F}$, respectively. Given a transform  $T$, the KL-divergence derived lower bound similarity $S$ is then given by,
\begin{equation}
\resizebox{.91\linewidth}{!}{$
    \displaystyle
    S(T(\mathcal{M}),\mathcal{F}) = \sup_{\Phi} E_{P_{T(\mathcal{M}),{\mathcal{F}}}}[\Phi] - \log E_{P_{T(\mathcal{M})} P_{\mathcal{F}}}[e^{\Phi}]
$} .
\end{equation}
By this means, an optimization over a mapping function $\Phi$ can be applied to compute the similarity of images,  instead of calculating the joint and marginal distributions of them.

We implement a CNN-based network, named the KLDivNet, to fit the function $\Phi$. 
As shown in Figure \ref{fig:Figure2}, the KLDivNet has a Y-shaped structure, 
which inputs a pair of images and outputs a final feature map $Z$.
Each input branch has two convolution blocks and a downsample layer, and the output branch has two convolution blocks.
The convolution block has a $1\times1\times1$ convolution layer and a leaky-ReLU activation layer, and no normalization layer is used. 
Each {\it locus} on the feature map, $Z_i$, represents the mapping result of correspondent {\it locus} on the image pairs $T(\mathcal{M})_i$ and $\mathcal{F}_i$, with $i$ denotes a patch pairs in the two images.

We denote KLDivNet as a function $\text{D}_\theta$, with $\theta$ denoting the parameters. 
To calculate $E_{P_{T(\mathcal{M}),{\mathcal{F}}}}[\Phi]$, we feed the network with the image pair ($\mathcal{F}$ and $T(\mathcal{M})$), and compute the mean of the feature map $Z = \text{D}_\theta(T(\mathcal{M}), \mathcal{F})$.
For $E_{P_{T(\mathcal{M})} P_{\mathcal{F}}}[e^{\Phi}]$, we adopt to shuffle the voxels in the fixed image, denoted as $\mathcal{F}^s$. Note that the shuffled image has the same marginal distribution as the original fixed image, but it is independent from the moved image. The similarity metric can be then computed,
\begin{align}
S(&T(\mathcal{M}), \mathcal{F}) =\nonumber \\
  &\sup_\theta {\text{mean}[\text{D}_\theta(T(\mathcal{M}), \mathcal{F})] – \log \text{mean}[e^{\text{D}_\theta(T(\mathcal{M}), \mathcal{F}^s)}]} .
\end{align}

\begin{figure*}[t]
  \centering
  \includegraphics[scale=0.57]{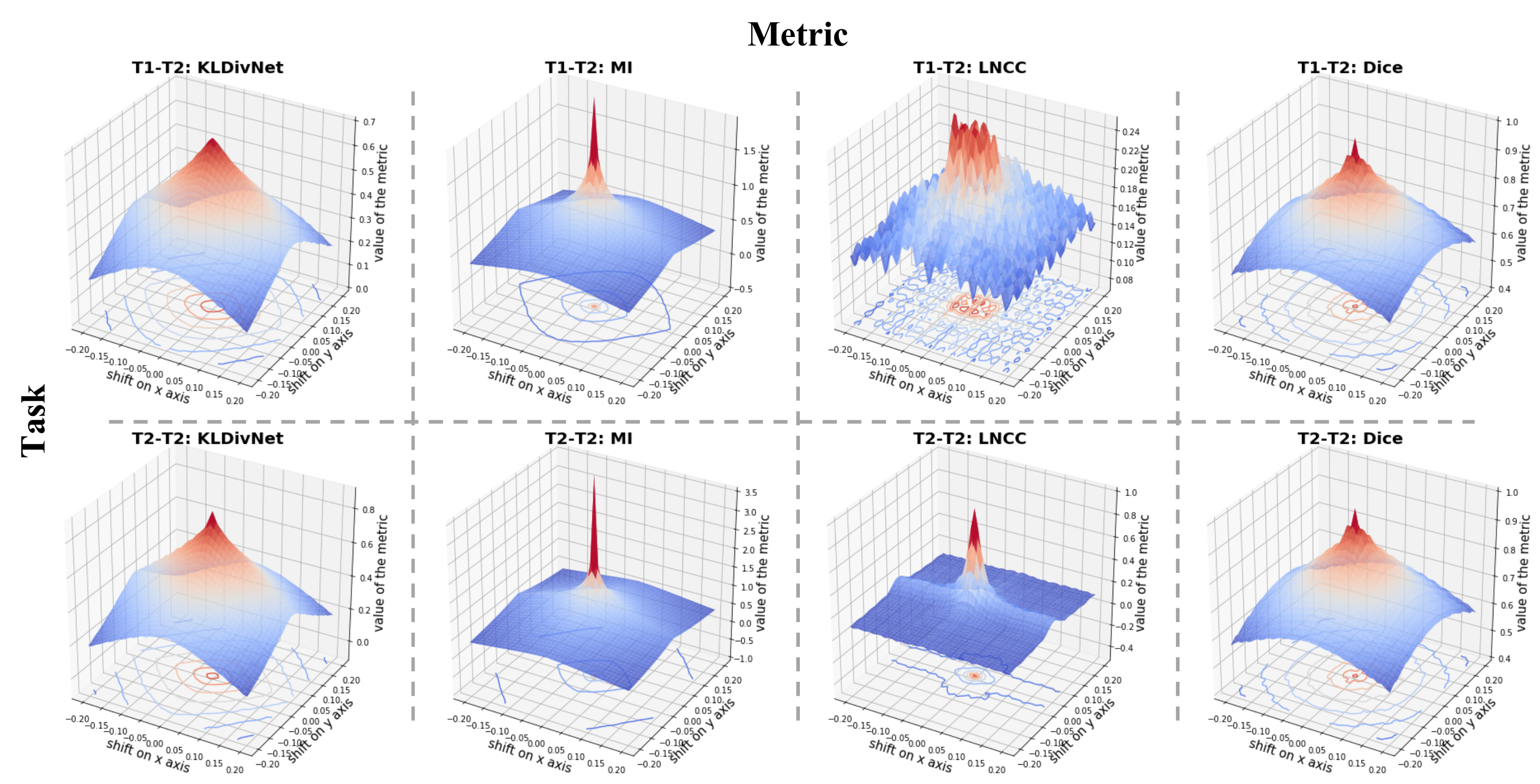}
  \caption{Plots of values of the similarity metrics for intra- and inter-modality images with different misalignments. The test images are from the AAL template, and the misalignments are different translation on x- and y-axis. For each plot, the contour of the metric is shown in the corresponding bottom plane.
}
  \label{fig:Figure3}
\end{figure*}

\subsection{The Registration Network}

Based on the KLDivNet, we implement an end-to-end multi-modality registration network, i.e., DivRegNet, as shown in Figure \ref{fig:Figure2}.
The main body, referred to as RegNet, follows the architecture of VoxelMorph \cite{VoxelMorph2018,VoxelMorph2019}, and has a Unet-like structure composed of four encoding blocks, four decoding blocks and skip connections. 
In decoding blocks, the scheme of SE Block \cite{SEBlock2018} is adopted to apply attention mechanism to channels. 
In both types of blocks, the switchable normalization \cite{SwitchableNorm} is used. 
In the training phase, both the RegNet and KLDivNet are trained via the maximization of the proposed similarity metric (or minimization of the derived loss); in the test (inference) phase, only the RegNet is needed.

\begin{table}[t]
  \centering
  \begin{tabular}{lll}
  \toprule
  Dataset  & Type & Count \& Modality \\
  \midrule
  AAL Brain     & \begin{tabular}[c]{@{}l@{}}Multi-modality,\\   Intersubject\end{tabular}  & \begin{tabular}[c]{@{}l@{}}1 T1, 1 T2, 1 PD,\\   250 Phantom T2\end{tabular}  \\
  LiTS Liver    & \begin{tabular}[c]{@{}l@{}}Single modality,\\   Intersubject\end{tabular} & 131 CT                           \\
  Hospital Liver& \begin{tabular}[c]{@{}l@{}}Multi-modality,\\   Intersubject\end{tabular}  & 14 CT, 19 MR                       \\
  \bottomrule
  \end{tabular}
  \caption{Summarizations of three datasets.}
  \label{tab:Table1}
\end{table}
\section{Experiments}

\subsection{Datasets and Metrics}

Three datasets were used to validate the proposed method, namely AAL Brain, LiTS Liver \cite{bilic2019liver} and Hospital Liver.
Table \ref{tab:Table1} summarizes their brief information.

The AAL Brain is a phantom dataset generated from the Anatomical Automatic Labeling (AAL) template \cite{rolls2019automated} of brain MR images. The AAL template contains a T1-MR image, a T2-MR image and a PD image.
For data augmentation, we generated 250 T2 images by using simulated deformation fields, which were obtained FFD transformations with 20mm isotropic spacings. The displacements of the control points in the FFD were random values from a gaussian distribution with zero mean and 6mm standard deviation.
The 250 phantom T2 images were then split into 150/50/50 ones, respectively for train/val/test sets, where the AAL template was chosen to be the fixed image, and the generated T2 images were the moving ones.

LiTS Liver is an open-source dataset initially for liver lesion segmentation of CT scans. A total of 131 CT images were split into three sets of 70/30/31.

Hospital Liver is a dataset of CT and MR scans from different patients collected from an anonymous hospital. Fourteen CT images were split into sets of 6/4/4, and 19 MR images were split into 10/4/5. The experiments only focused on the deformable registration, thus the preprocessing procedures including linear-registration and normalization were applied. Random translation and rotation on both fixed and moving images were used for data augmentation during the training phase.

Similar to the registration network with KLDivNet loss, we implemented it with the LNCC loss, the autograd-compatible MI loss based on Parzen window density estimation (PWDE) with gaussian kernel \cite{sandkuhler2018airlab}, for comparisons.


For the training, the ADAM optimizer with 0.001 learning rate was used, and a typical training schedule of 10,000 iterations was adopted. Besides, exponential moving average (EMA) was used to ensure a stable performance in validation and test phases.

To evaluate the performance of the networks, we used three different metrics, namely the Dice coefficient, average surface distance (ASD) and Hausdorff Distance (HD). 

\begin{table*}[t]
\begin{tabular}{l|rrr|rrr|rrr}
\toprule
\multirow{2}{*}{Method} & \multicolumn{3}{c|}{T1-T2}  & \multicolumn{3}{c|}{PD-T2} & \multicolumn{3}{c}{T2-T2} \\
& \multicolumn{1}{c}{\small{$\uparrow$Dice}} & \multicolumn{1}{c}{\small{$\downarrow$ASD}\scriptsize{(mm)}} & \multicolumn{1}{c|}{\small{$\downarrow$HD}\scriptsize{(mm)}} & \multicolumn{1}{c}{\small{$\uparrow$Dice}} & \multicolumn{1}{c}{\small{$\downarrow$ASD}\scriptsize{(mm)}} & \multicolumn{1}{c|}{\small{$\downarrow$HD}\scriptsize{(mm)}} & \multicolumn{1}{c}{\small{$\uparrow$Dice}} & \multicolumn{1}{c}{\small{$\downarrow$ASD}\scriptsize{(mm)}} & \multicolumn{1}{c}{\small{$\downarrow$HD}\scriptsize{(mm)}} \\
\hline
Affine                  & 0.9039                             & 1.34                                    & 9.84                                   & 0.9039                             & 1.34                                    & 9.84                                   & 0.9039                             & 1.34                                    & 9.84                                   \\
FFD SEMI                & 0.9288                             & 1.06                                    & \textbf{9.14}                                   & 0.9281                             & 1.04                                    & \textbf{9.04}                                   & 0.9293                             & 1.04                                    & 9.47                                   \\
RegNet + LNCC           & 0.8709                             & 1.52                                    & 9.90                                   & 0.9164                             & 1.18                                    & 9.50                                   & 0.9499                             & 0.80                                    & 9.47                                   \\
RegNet + MI {[}a{]}     & 0.9296                             & 1.02                                    & 9.84                                   & 0.9306                             & 1.01                                    & 9.71                                   & 0.9477                             & 0.82                                    & 9.52                                   \\
RegNet + MI {[}b{]}                 & 0.9524                             & 0.75                                    & 9.70                                   & 0.9484                             & 0.80                                    & 9.71                                   & 0.9575                             & 0.69                                    & 9.62                                   \\
RegNet + MI {[}c{]}                 & 0.9527                             & 0.75                                    & 9.78                                   & 0.9491                             & \textbf{0.78}                                   & 9.74                                   & 0.9575                             & 0.69                                    & 9.46                                   \\
DivRegNet (ours)    & \textbf{0.9546}                             & \textbf{0.72}                                    & 9.60                                   & \textbf{0.9501}                             & \textbf{0.78}                                    & 9.73                                   & \textbf{0.9585}                             & \textbf{0.67}                                    & \textbf{9.41}                                  \\

\bottomrule
\end{tabular} \\
\footnotesize{
Our implementation of MI may be affected by the number of bins when estimating distribution. {[}a{]}: MI registration with the number of histogram bins setting to be 16; MI {[}b{]}: the number of bins is 64; MI {[}c{]}: the number is 256.
}
\caption{The performance of different methods on AAL Brain dataset.}
\label{tab:Table2}
\end{table*}

\begin{figure}[t]
  \centering
  \includegraphics[scale=0.6]{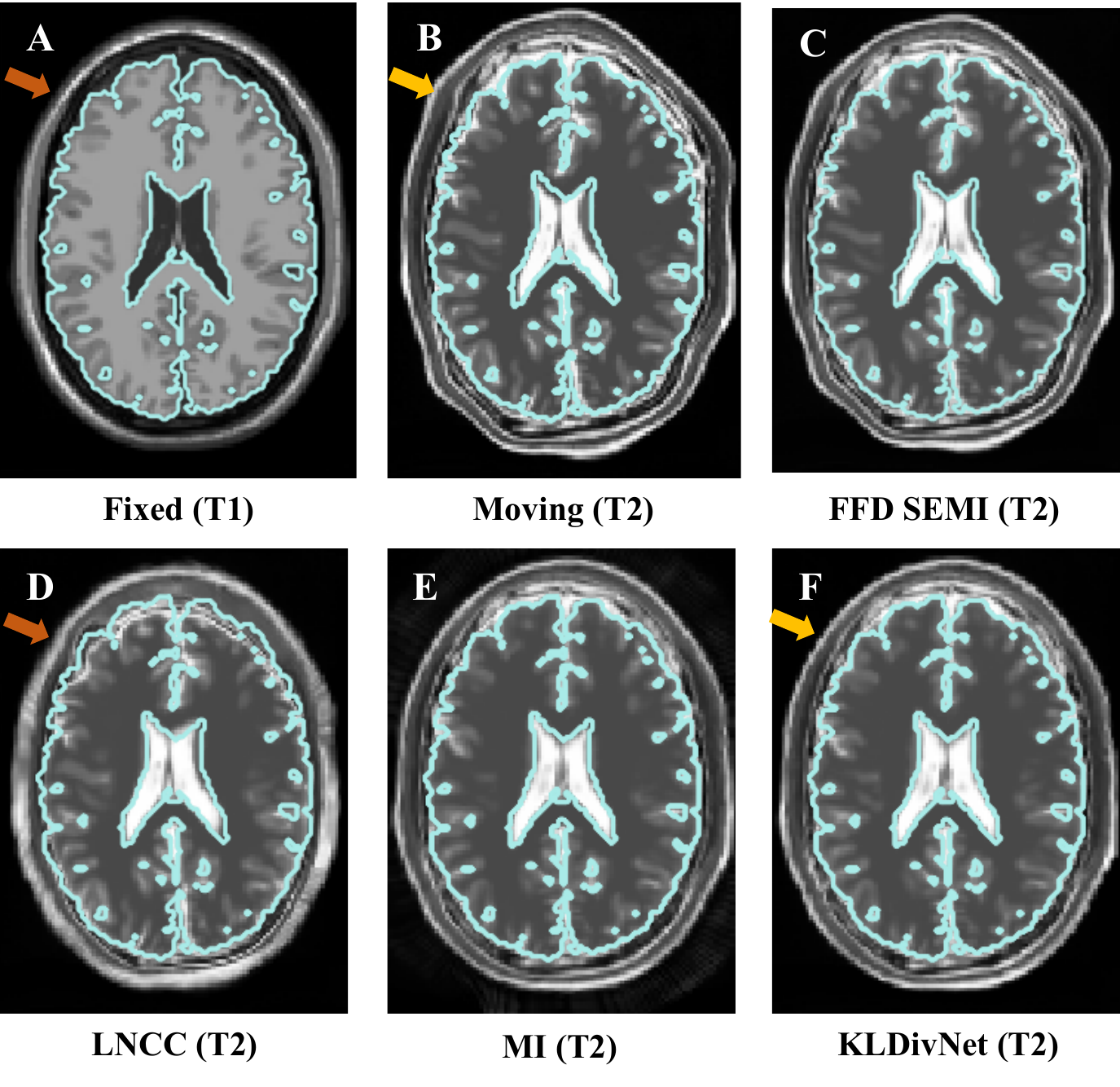}
  \caption{Visualization of the registration results in AAL Brain T1-T2 task. Cyan contours indicate the label of cinereum matter on the fixed image. Orange arrows mark T1-like feature of the skull, and yellows mark T2.}
  \label{fig:Figure4}
\end{figure}

\subsection{Effectiveness of the proposed similarity metric}\label{exp:effect}

In this study, we examine the effectiveness of similarity metric by KLDivNet. For comparisons, the study includes the other similarity metrics as loss functions for both intra- and inter-modality images.

We employed T1-T2 and T2-T2 images of the AAL Brain dataset for illustration. Random misalignments of translations were applied to the paired images, and values of the three metrics, as well as the Dice coefficient of cinereum matter, were computed and plotted in Figure~\ref{fig:Figure3}.
As shown in the figure, in the single-modality T2-T2 image registration, all three similarity metrics perform well.
By contrast, in the inter-modality T1-T2 case the LNCC shows a large number of local optimum, indicating it is not suitable for this cross-modality image registration task. 
Both the KLDivNet and MI similarity metrics demonstrate perfect global optima, similar to that of the Dice scores. 
This study confirms that the proposed similarity metric from KLDivNet is suitable for multi-modality image registration as well as for single-modality image registration. 


\subsection{Registration performance and comparisons}

This study investigates the registration performance of the proposed DivRegNet and compares with the other registration methods, including the FFD SEMI registration \cite{journal/tmi/Zhuang11}, the RegNet with LNCC loss and the RegNet with MI loss.

For FFD SEMI which is a conventional iterative optimization registration method, we used the online public software provided by Zhuang {\it et al.}\shortcite{journal/tmi/Zhuang11}\footnote{http://www.sdspeople.fudan.edu.cn/zhuangxiahai/0/zxhproj/}, and the registration was run with default parameters.
For the RegNet with MI loss, since we used the autograd-compatible MI based on PWDE, we needed to manual set the number of histogram bins used in the estimation of joint probability distribution of image intensity, which can be critical for the accuracy of registration. Therefore, we first used three values for the number, i.e., 16, 64 and 256, in the AAL Brain experiment. 


\subsubsection{Registration for AAL Brain}
This study includes two multi-modality tasks, i.e., T1-T2 and PD-T2 registration, and one single-modality task, T2-T2.
The results are given in Table \ref{tab:Table2}, where the accuracies after affine registration are also provided for reference. 

In the single-modality (T2-T2) registration task,  all of the deformable registration methods performed well. One can see that the compared methods generated different registration accuracies, and the proposed DivRegNet obtained the best figures in all the three evaluation categories.

In the multi-modality registration tasks, including T1-T2 and PD-T2 image registration, RegNet+LNCC did not perform as well as the other methods, indicating LNCC is not suitable for cross-modality image registration. This is consistent to results Section~\ref{exp:effect}.
For the other methods, one can see that DivRegNet performed the best in terms of Dice and ASD. However, the conventional FFD+SEMI obtained the best HD values, indicating the deformation fields generated by FFD+SEMI could be more realistic, even though its accuracy was limited due to the complex iterative optimization procedure.

For the RegNet+MI registration, the three implementations all performed well in all the three tasks.
However, the performance was indeed affected by the different setting of the histogram estimation. 
RegNet+MI with 16 histogram bins was evidently worse than the other implementations. 
This is probably due to its less accurate estimation of the intensity distributions. 
The greater numbers used the better performed RegNet+MI demonstrated. 
However, the difference between MI with $bins=64$ and MI with $bins=256$ become very small. 
Furthermore, the computation time consumed by MI with $bins=256$ is four times of that of MI with $bins=64$ in computing the marginal distributions and 16 times in computing the joint distribution.  

Figure \ref{fig:Figure4}  visualizes the registration results of a typical case in multi-modality registration. 
As the figure shows, both  RegNet+MI and  KLDivNet provided good registration. By contrast, RegNet+LNCC misaligned the image by shrinking the cortical area. 
Moreover, as pointed out by the orange arrow in Figure \ref{fig:Figure4}(A), the skull in T1 images has a dark-light-dark feature, while the skull in T2  has a light-dark-light feature, pointed out by the arrow in Figure \ref{fig:Figure4}(B). When using the LNCC loss, the RegNet tended to wrap the skull to match T1-like feature, which is erroneous in clinics.

\begin{table}[t]
{\begin{tabular}[l]{@{}lrrr}
    \toprule
      Method & \multicolumn{1}{c}{$\uparrow$Dice} & \multicolumn{1}{c}{$\downarrow$ASD(mm)} & \multicolumn{1}{c}{$\downarrow$HD(mm)} \\
      \midrule
        Affine                  & 0.7682                             & 8.02                                    & 51.06                                  \\
      FFD SEMI                & 0.7905                             & 7.34                                    & 49.22                                  \\
      VoxelMorph{[}a{]}     & 0.831{[}c{]}                       & -                                       & -                                      \\
      1-cascade RCN{[}b{]}   & 0.867{[}c{]}                       & -                                       & -                                      \\
      RegNet + LNCC           & 0.8777                             & 5.86                                    & 45.53                                  \\
      RegNet + MI {[}d{]}     & 0.9016                             & 4.98                                    & \textbf{42.37}                                  \\
      DivRegNet (ours)    & \textbf{0.9030}                            & \textbf{4.97}                  & 42.69             \\
      \bottomrule
    \end{tabular}}\\
    \footnotesize{{[}a{]}: Balakrishnan {\it et al.} \shortcite{VoxelMorph2019}}\\
    \footnotesize{{[}b{]}: Zhao {\it et al.} \shortcite{Recursivecascaded2019}.}\\
    \footnotesize{{[}c{]}: Both results are reported by Zhao {\it et al.} \shortcite{Recursivecascaded2019}.}\\
    \footnotesize{{[}d{]}: MI with number of histogram bins being 64.} 
    \caption{The Performance on LiTS Liver Dataset.}
    \label{tab:Table3}
\end{table}

\begin{table}[t]
  \begin{tabular}[l]{@{}lrrr}
  \toprule
  Method               & \multicolumn{1}{c}{$\uparrow$Dice} & \multicolumn{1}{c}{$\downarrow$ASD(mm)} & \multicolumn{1}{c}{$\downarrow$HD(mm)} \\
  \midrule
  Affine               & 0.8084                             & 5.35                                    & 44.29                                  \\
  FFD SEMI             & 0.8304                             & 4.81                                    & \textbf{43.73}                                  \\
  RegNet + LNCC        & 0.8247                             & 5.12                                    & 43.11                                  \\
  RegNet + MI {[}a{]}  & 0.8348                             & 4.77                                    & 44.09                                  \\
  DivRegNet (ours) & \textbf{0.8351}                             & \textbf{4.72}                                   & 43.88                                   \\
  \bottomrule
  \end{tabular}\\
  \footnotesize{{[}a{]}: MI with number of histogram bins being 64.} 
  \caption{The Performance on Hospital Liver Dataset.}
  \label{tab:Table4}
\end{table}

\subsubsection{Registration for LiTS and Hospital Liver}

Alongside with experiments on brain images, we also examine three losses on two liver datasets, i.e., the LiTS Liver dataset for single modality registration and the Hospital Liver dataset for multi-modality registration. For the LiTS Liver dataset, the fixed and moving image pairs were random sampled from train/val/test subsets, for example, for test, there are in total 930 pairs. For the Hospital Liver dataset, when generating data pairs, one of the CT image was sampled to be the fixed image, and one of the MR to be the moving one. 

As listed in Table \ref{tab:Table3}, all the methods performed well on the LiTS Liver dataset. Results of the same dataset from VoxelMorph \shortcite{VoxelMorph2019} and Recursive Cascaded Networks (RCN) \shortcite{Recursivecascaded2019} are also listed. For RCN, only the 1-cascade variant, i.e. the integrated affine and one-stage deformation registration network is used for reference. For the RegNet+MI, according to the experiments on AAL Brain dataset, we chose $bins=64$ as a balance of performance and the computation time.  
As is shown, the DivRegNet obtained best values in terms of Dice and ASD, while the RegNet+MI performed best in HD. The DivRegNet was inferior to the RegNet+MI in terms of HD, but the interval between these two methods was small. The results of the RegNet+LNCC, though were relatively lower than other two methods, were still comparable to the reported results. Hence, the performance of the DivRegNet is promising over single modality tasks. 

As for multi-modality task, performance on Hospital Liver Dataset is demonstrated in Table \ref{tab:Table4}. The Dice of Affine registration reached 0.8, showing the difference among subjects is quite small. In this task, the MI based method, i.e., RegNet+MI and FFD SEMI, outperformed the RegNet+LNCC, but their performances still fell a little below our DivRegNet. The performance of FFD SEMI was consist with its performance in AAL Brain, with a probable cause that the smoothness of FFD is suitable for simulating the truly deformation. The competitive performance of the DivRegNet shows its great potential in multi-modality registration.

\section{Conclusion}
In this work, we propose a new image similarity metric for multi-modality image registration. 
The metric is derived from the lower bound of the KL-divergence, based on the Donsker-Varadhan representation, and is implemented using a DNN, thus referred to as KLDivNet. 
With the KLDivNet, we implement a multi-modality registration framework, i.e., DivRegNet.
DivRegNet can be trained in an unsupervised manner, thanks to the advantage of the intensity-based similarity metric provided by KLDivNet which can also be trained unsupervisedly. 
We illustrated KLDivNet being suitable for both the intra- and inter-modality image registration.
We validated DivRegNet using three datasets for different clinical applications. Results shows that the DivRegNet was capable for multi-modality registration and delivered state-of-the-art performance for registration.

\bibliographystyle{named}
\bibliography{ijcai20}

\end{document}